# A Secure Healthcare 5.0 System Based on Blockchain Technology Entangled with Federated Learning Technique

Abdur Rehman[a], Sagheer Abbas[a], M. A. Khan[b], Taher M. Ghazal[c,d], Khan Muhammad Adnan[e*], Amir Mosavi[f,g,h*]

[a] *School of Computer Science, national College of Business Administration and Economics, Lahore, 54000, Pakistan; (S.A), (A.R)*
[b] *Riphah School of Computing and Innovation, Faculty of Computing, Riphah International university, Pakistan; (M.A.K)*
[c] *School of Information Technology, Skyline University College, University City Sharjah, 1797, Sharjah, UAE, (T.M.G)*
[d] *Center for Cyber Security, Faculty of Information Science and Technology, Universiti Kebangsaan Malaysia (UKM), 43600, Bangi, Selangor, Malaysia*
[e] *Department of Software, Gachon University, Seongnam, 13120, Republic of Korea (K.M.A)*
[f] *Institute of Information Engineering, Automation and Mathematics, Slovak University of Technology in Bratislava, 81107 Bratislava, Slovakia; amirhosein.mosavi@stuba.sk (A.M)*
[g] *John von Neumann Faculty of Informatics, Obuda University, 1034 Budapest, Hungary;*
[h] *Faculty of Civil Engineering, TU-Dresden, 01062 Dresden, Germany;*

**Abstract:** In recent years, the global Internet of Medical Things (IoMT) industry has evolved at a tremendous speed. Security and privacy are key concerns on the IoMT, owing to the huge scale and deployment of IoMT networks. Machine learning (ML) and blockchain (BC) technologies have significantly enhanced the capabilities and facilities of healthcare 5.0, spawning a new area known as "Smart Healthcare." By identifying concerns early, a smart healthcare system can help avoid long-term damage. This will enhance the quality of life for patients while reducing their stress and healthcare costs. The IoMT enables a range of functionalities in the field of information technology, one of which is smart and interactive health care. However, combining medical data into a single storage location to train a powerful machine learning model raises concerns about privacy, ownership, and compliance with greater concentration. Federated learning (FL) overcomes the preceding difficulties by utilizing a centralized aggregate server to disseminate a global learning model. Simultaneously, the local participant keeps control of patient information, assuring data confidentiality and security. This article conducts a comprehensive analysis of the findings on blockchain technology entangled with federated learning in healthcare. 5.0. The purpose of this study is to construct a secure health monitoring system in healthcare 5.0 by utilizing a blockchain technology and Intrusion Detection System (IDS) to detect any malicious activity in a healthcare network and enables physicians to monitor patients through medical sensors and take necessary measures periodically by predicting diseases. The proposed system demonstrates that the approach is optimized effectively for healthcare monitoring. In contrast, the proposed healthcare 5.0 system entangled with FL Approach achieves 93.22% accuracy for disease prediction, and the proposed RTS-DELM-based secure healthcare 5.0 system achieves 96.18% accuracy for the estimation of intrusion detection.

## 1. Introduction

The IoMT is an emerging technology that is rapidly expanding [1]. The Internet of Things (IoT) enables the connection of numerous objects to collect data that may be used to enhance human health, productivity, and effectiveness [2-4]. Smart cities, smart grids, and smart houses are well-established concepts that are transforming our daily lives [5-7]. Among the most potential new technical approaches for addressing the global health equality gap is the use of an IoT-based monitoring system of patient health [8]. Another name for these IoT technologies is the IoMT. In this study, the terms IoT and IoMT are used interchangeably, even though the study will focus on the healthcare industry. Furthermore, the Internet of Things has the potential to improve healthcare and public safety significantly [9]. Individuals can obtain information on their lifestyles, physical and mental efficiency, and living surroundings, among other things, by connecting their beings to the Internet. This allows healthcare providers to monitor people's health remotely and in real-time. Furthermore, the data gathered could be used to promote evidence-based interventions for diseases, trauma, protection, early detection, and rehabilitation. In today's world, transferring patients from their homes to hospitals for routine check-ups is extremely difficult. There are several challenges, including queuing, travel time, and the possibility of patients contracting various viruses while traveling through this polluted environment. As a result, the healthcare industry is focusing on in-home healthcare services, which allow patients to conduct medical examinations in the comfort of their own homes. A smart health monitoring system is designed to help patients who live in remote areas contact doctors in urban areas. This technology acts as a bridge between patients and clinicians. It keeps track of vital signs like heart rate, electrocardiogram (ECG), blood pressure, temperature, and whether or not a person has fallen. The system collects this data and sends it to the application for further analysis via a wireless connection. Effective mining algorithms are urgently needed to analyse medical information to help in disease discovery, offer medical treatment, and enhance patient care. Machine learning is a sophisticated computational technique that has been used in a variety of domains such as image recognition, language processing, and health care [10]. Nonetheless, machine learning models acquire great accuracy only with a vast quantity of training set, which is crucial in healthcare, where precision may sometimes mean the difference between saving or losing a patient's life. In most cases, centralized training strategies include acquiring a big volume of information from a robust cloud server, which might lead to major consumer privacy violations, especially in the medical field. As an open and accountable data protection mechanism, the development of blockchain technology paves the way for new ways to address key issues of privacy, security, and ethics in a smart healthcare system. However, blockchain has achieved excellent success as a backbone of cybersecurity architecture for a variety of smart healthcare technologies, such as the control of patient record access, information distribution, etc. [11]. It is justifiable to incorporate blockchain in intelligent domestic networks as it is autonomous of heterogeneous protocols that are often used in smart systems [12]. However, notwithstanding increased interest in smart healthcare technology, current work is distributed across diverse fields of study. Such a timely study is also being undertaken to close the void and have practical insights on blockchain technology, techniques, and their application in the field of the intelligent healthcare system. The IoMT has generated security flaws in the healthcare system. Unified networks enable hackers to reach the linked devices for malicious purposes. The big issue is hack attempts on medical records and healthcare devices; not only are they the key components of a smart healthcare system but they can also be used for malicious purposes: delivering phishing and spam mail. Because of unencrypted wireless keys, smart healthcare devices are often the central target for DDoS attacks, particularly because they are immediately turned on to provide smarter solutions such as patient medical records and the automatic updating of false medical reports [13]. These challenges originate from a centralized IoMT system structure, and privacy issues are growing as the IoMT revolution progresses, namely record forgery and manipulation, device interference, and infiltration of illegal IoMT devices via attacks against server and gateway networks [14].

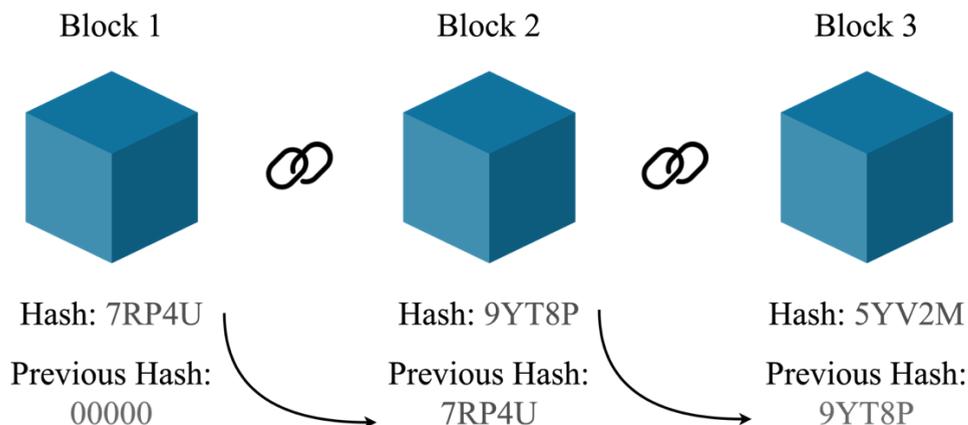

**Figure 1:** Blockchain Structure

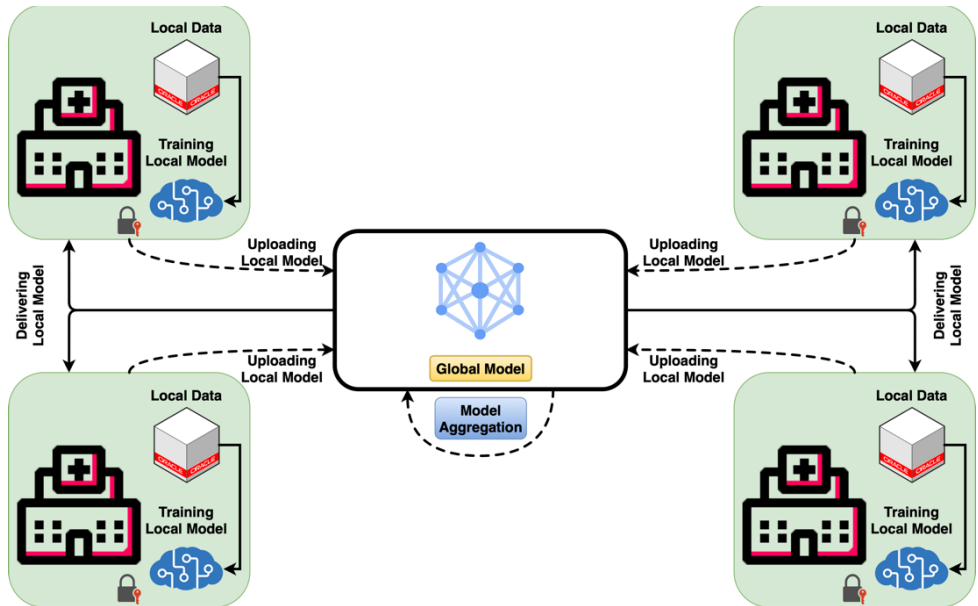

**Figure 2:** A Federated Learning Technique

These issues can be addressed through the deployment of blockchain-based systems and unified "cloud-like" computer networks [15]. Satoshi Nakamoto created the blockchain technology in 2008, which featured a time-stamped collection of harmful proof documents that was maintained by a network of independent networks. Blockchain architecture is illustrated in Figure 1. It is a series of blocks that are connected with simple cryptography. Inflexibility, decentralization, and transparency are the three main concepts in the functioning of Blockchain technologies. The three responsibilities have been highly productive and opened their doors to a broad array of technologies relevant to the virtual currency, such as the functioning of autonomous vehicles, such as smartphones, and embedded devices. Whereas Blockchain technology is safe and enables anonymity, there are still certain drawbacks to it in the current stage of deployment [16]. FL is a decentralized machine learning platform for the IoT that enables numerous devices to collaboratively learn machine learning models without exchanging their actual data. Figure 2 depicts FL's architecture. This enhances the intelligent healthcare system by avoiding the leaking of patient information. Figure 2 shows a federated learning-based healthcare system in which embedded sensors gather medical information from healthcare providers, multiple edge devices collaborate on federated learning algorithms, and machine learning techniques assess the patient's well-being and, if needed, seek immediate assistance in the cloud. Federated learning is a recently popular paradigm due to the incredible assurance it provides for studying fragmented sensitive information. Rather than combining data from disparate sources or relying on the traditional find-then-replicate strategy, it enables training of a common global model on a centralized server while retaining data in the relevant organizations. This method, known as federated learning, allows individual locations to work together to train a global model. Federated learning is the process of pooling training data from multiple sources to develop a global model without directly exchanging datasets. This ensures that patient privacy is maintained across sites.

Healthcare 5.0 [17-19] is a network architecture that requires fifth-generation communication as the core network architecture for linking healthcare equipment. IoT will create data that AI can utilize, contributing to the advancement of digital wellbeing by concentrating not just on patient well-being and quality of life, but also on the well-being and quality of life of people worldwide. The primary issues in healthcare 4.0 are flawless information transfer with little or no information leakage. Automation and AI are two emerging healthcare 5.0 technologies that have the potential to transform any type of employment. Intelligence in healthcare 5.0 encompasses ideas such as accurate and systematic disease diagnosis, virtual patient monitoring and detection, remote surgery, and intelligent treatment, that comprises online training for patients suffering from anxiety. Artificial Intelligence (AI) is a broad concept that encompasses all intelligent advancements. It refers to the ability of machine-learning methods to predict final results without the involvement of humans. As a result of the tremendous expansion and advancement of technology in healthcare, the term "smart health system" has been coined. In light of these capabilities, numerous solutions have been proposed for use in numerous industries, including smart homes, Industrial Internet of Things (IoT), and smart healthcare. However, as privacy threats become increasingly sophisticated, there are still a number of obstacles to overcome when implementing blockchain-based federated learning in healthcare 5.0:

1. The in-blockchain features of the model can potentially be used by attackers to determine the original secret healthcare data.
2. Some clinical data from medical devices may be falsified to deceive the FL process.
3. There is no motivation for medical devices to provide data and processing power to FL.

To address the aforementioned problems, this study blends FL and sophisticated encryption to provide a safe and privacy-preserving healthcare 5.0 system. The following are the most significant contributions made by this study;

1. We present a blockchain-based FL framework for healthcare 5.0 that not only builds an accurate multidisciplinary model based on multiple edge devices, but also supervises the whole training process.
2. The proposed method provides an additional degree of security to blockchain-based FL, we present an RTS-DELM method that balances privacy and model accuracy by adjusting noise depending to the training process.
3. The proposed system considers multiple medical organizations in the proposed model because the locally trained model of other medical organizations may enhance the capability of the healthcare 5.0 system by sharing a global model.
4. To implemented the federated learning approach in healthcare 5.0 to improve the learning process of clinical data by locally training models
5. To provide an intelligent hybrid approach to enhance secure communication and effective healthcare monitoring.
6. We design an Intrusion Detection system (IDS) in healthcare 5.0 system that enhances the security and privacy by detecting intrusions and attack patterns.

The proposed method secures the FL-based healthcare 5.0 Network by accurately assessing its reliability with regard to the most important security goals of secrecy, validity, and accessibility. To support the claim that the overhead caused by the proposed technique is minimal compared to the value of its security and privacy, the study assesses the suggested method's capacity to securely protect sensitive data while consuming quite minimal resources. The purpose of this study is to examine a system model based on a Real-Time Deep Extreme Learning System (RTS-DELM) for intelligent disease prediction and intrusion detection in healthcare 5.0 that achieve the maximum possible level of accuracy. The remaining portions of this research study can be broken down into the following categories. Section 2 offers an overview of relevant research. Section 3 enlightens the proposed methodology. Section 4 presents the simulation and findings of the suggested method. Finally, section 5 discusses the study's findings.

## 2. Literature Review

The term "blockchain" has been popular among proponents of smart healthcare in recent years, and several research publications have explored how blockchain technology may be applied in the field. S. Aggarwal et al. [20] explored several facets of healthcare, such as the integration of transactions, home healthcare, and investment distribution. The smart home industry has several potential applications for the blockchain. M. Andoni et al. [21] offered a comprehensive analysis of the many blockchain applications of a P2P resource sharing network. The report provides in-depth knowledge on the deployment and abilities of several smart home networks, including smart grid security issues, Big Data analysis, AI and payment services. They concluded that the study did not sufficiently take into consideration challenges associated with smart homes, such as financial planning for smart cities and smart home security. G. Li et al. [22] suggested a blockchain structure that would be based on users to ensure the safety of information communication in the IoT. Z. Zhou et al. [23] conducted research into various blockchain techniques, predetermined investigation, and decentralized computing to relocation control over particular automobiles and enhance their effectiveness. Du et al. [24] proposed a study with the goals of investigating the implementation of blockchain technology in smart healthcare, developing a centralized conceptual approach for smart healthcare, outlining the impact of blockchain technology on advanced health care system, and eventually, developing a stakeholder-based advancement application framework for smart healthcare. Ihnaini et al. [25] suggested a smart diabetic disease prediction method centered on deep machine learning and information fusion concepts. By combining information, the proposed technique can reduce unnecessary strain on the system's computing resources while also improving the proposed system's efficiency in correctly predicting and recommending this life-threatening condition. Finally, an ensemble machine learning approach is used to create a diabetes prediction model. Nowadays, data can be easily exchanged across multiple networks, allowing experts and organizations to make the best use of existing capabilities while meeting society's medical demands. Users can gain access to strong and efficient healthcare services thanks to the Internet of Things. The use of smart sensors has aided in the proper monitoring of community healthcare demands. Wearable devices can be used to monitor a wide range of bodily functions. Some can be integrated to monitor various bodily systems to ensure that medical services are delivered to such people in a beneficial manner. The data collected in this manner can be examined, pooled, and mined to perform effective disease prediction [26]. Khan et al. [27] suggested novel healthcare facilities for senior citizens focused on the patients' actual needs and problems. To better satisfy the basic demands of elderly healthcare, the researchers applied machine learning approaches.

Xu et al. [28] presented an overview of federated learning techniques, focusing on those used in biomedicine. Review and explain the broad solutions to the statistical challenges, system challenges, and privacy concerns inherent in federated learning, while emphasizing the implications and potential for healthcare. Li et al. [29] offered an overview of the implementation of ML and bioinformatics technologies in the medical business utilizing bibliometric visualization and Web of Science (WOS). A review focuses on the countries that conduct the most research, the primary research subjects, funding sources, and hotspots for research in this field. In addition, the study outlines major difficulties and future research objectives for the use of machine learning and deep learning methods in the healthcare industry. Siddiqui et al. [30] applied the data fusion technique in a deep learning model to predict breast cancer stages. They applied decision-based fusion to increase the accuracy of the suggested methodology. Medjahed et al. [31] proposed an intelligent healthcare monitoring system based on a data fusion approach. The proposed system is based on a multi-sensor platform that can enable full control over smart homes. Using AI applications to help diagnose diseases and improve disease prognosis and minimize patient treatment times has become standard practice since the development of artificial intelligence. Dai et al. [32] transformed the challenge of hospitalization forecasting into a supervised classification issue, leading to a significant set of possible medical expense savings. Son et al. [33] used a Support Vector Machine (SVM) algorithm to determine drug compliance in individuals with heart problems. Tariq et al. [34] created a heterogeneous fusion Artificial intelligence-based method to forecast the intensity of COVID-19 using historical medical information. Sedik et al. [35] proposed a methodology for feature extraction to enlarge the information and employed convolutional neural networks and convolutional long short-term memory algorithms to identify coronavirus. Qayyum et al. [36] suggested a clustered FL-based technique for processing medical visual data at the edge, enabling remote infirmaries to capitalize from multimodal information while maintaining their confidentiality. Brisimi et al. [37] forecasted timely treatments for patients with cardiovascular illnesses by utilizing FL to solve dispersed sparse Support Vector Machine difficulties. Nonetheless, the above-mentioned centralized training approaches necessitate the collection of confidential medical information in a unified databank, which is problematic owing to data security issues. Rather, FL appears as a decentralized architecture that enables cooperative knowledge while retaining all delicate information locally, delivering a private remedy for connecting disparate clinical data on end devices. Chang et al. [38] propose a blockchain-based federated learning approach for smart healthcare where the MIoT devices apply the federated learning to fully use the dispersed healthcare information and the edge nodes sustain the blockchain to avoid a data loss. In recent years, several research using federated learning in intelligent healthcare have been published as shown in table 1.

Table 1: Comparison of Literature with Proposed Model

| Authors/Objectives | Type of Data | Predictive Model | Decision Making | Fused Decision Making | Healthcare 5.0 Paradigm | Use of Blockchain | Use of IDS | Use of FL |
|---|---|---|---|---|---|---|---|---|
| Du et al. [24] | Medical Records | Yes | Yes | No | No | Yes | No | No |
| Medjahed et al. [31] | Medical Records | Yes | Yes | Yes | No | No | No | No |
| Ihnaini et al. [25] | Medical Records | Yes | Yes | Yes | No | No | No | No |
| Khan et al. [27] | Medical Records | Yes | Yes | No | No | No | No | No |
| Li et al. [29] | Medical Records | No | No | No | No | Yes | No | No |
| Siddiqui et al. [30] | Medical Records | Yes | Yes | Yes | No | No | No | No |
| Chang et al. [38] | Medical Records | Yes | Yes | No | No | Yes | No | Yes |
| **Proposed Model** | **Sensor data** | **Yes** | **Yes** | **Yes** | **Yes** | **Yes** | **Yes** | **Yes** |

## 3. Proposed Methodology

### 3.1. Blockchain Module Implementation

A blockchain was initially designed about and developed by Satoshi Nakamoto in the year 2008 [39]. The block comprises a vast amount of transaction information, including the block id, a hash of the previous block, transaction details, nonce, and time stamps. In a blockchain scheme, where miners determine the correct hash to add a block, the winners would first search the existing block before searching for a new block. The proof of work methodology is used to verify whether or not a given block of transactions is legitimate. The steps listed below describe the core components of blockchain technology. In a smart healthcare system, any node that is connected to the Internet must communicate with a repository of storage data, along with miners in a blockchain framework. The blockchain holds all unprocessed transactions pending an opening to a new block to validate it. Many of the transactions are first reviewed and then shortly analyzed by the Merkle tree. The new smart healthcare system connectivity ecosystem will be created by blockchain technology, as it is versatile and compliant with healthcare IoMT applications.

### 3.2. RTS-DELM Module Implementation

RTS-DELM is a data analytics platform that automates data analysis tasks and provides insightful information. A RTS-DELM approach analyses real-time data using the Deep Extreme Learning Machine (DELM). The DELM can be used to determine energy consumption levels, inventory services, and specify transportation operations, to name a few applications [40]. The RTS-DELM method can be used to modify datasets in healthcare networks, implying that any errors can be excluded. It employs a novel approach to disease prediction and diagnosis. The goal of this study is to evaluate a system model based on the RTS-DELM for the most accurate adaptive forecasting of healthcare monitoring.

### 3.3. Federated Learning Module Implementation

Federated learning is a recently popular paradigm due to the incredible assurance it provides for learning with fragmented sensitive data. Rather than combining data from multiple sources or relying on the traditional find-then-replicate strategy, it allows for the training of a common global model on a central server while retaining data in the appropriate organizations. Federated learning, as it is known, allows individual sites to contribute to the training of a global model. Federated learning is the process of combining training data from multiple sources to build a global model without directly sharing datasets. This ensures that patient privacy is preserved across locations.

Several firms or institutions collaborate to solve a machine-learning problem using federated learning, which is managed by a central server or service provider. As a result, a deep learning model is hosted and optimized by a central server. The model is trained by dispersing itself across remote centralized datacentres, which may include hospitals or other medical organizations, preserving data localization at these locations. Throughout the training process, no data from any participant is exchanged or transferred. Rather than sending data to a single server, as in traditional deep learning, the server maintains a globally shared architecture that is accessible to all institutions. Following that, each organization develops its model based on patient data. Following that, each institute communicates with the server using the error gradient of the model. The central server collects all participant feedback and modifies the global model according to predefined criteria. The predefined criteria allow the model to evaluate the excellence of the response and thus include only information that adds value. As a result, feedback from institutions reporting poor or unusual results may be overlooked. This method creates a single federated learning cycle that is repeated until the global model is acquired.

*3.4. Data Fusion Module Implementation*

Data fusion techniques combine information from multiple sensors to produce more precise observations than a single, independent sensor could. The process of obtaining data from disparate but likely linked sources and combining it to have the greatest impact is known as information extraction. A security framework includes several network security sensors, and it is challenging to gain a comprehensive, wide-angle perspective of the security system's dynamic security situation. Additionally, equipment spread across a vast region may be challenging to handle effectively. In order to enhance the model's efficacy and give a comprehensive analysis of the system's protection situation, it is also essential to efficiently and smartly combine the sensor data. In comparison, because the information comes from multiple sources, multiple data sources can provide a higher level of consistency in terms of trustworthiness.

Sensors contribute to the Internet of Things by serving as critical tools for evaluating smart healthcare systems, ecosystems, and consumers. The following devices fall under this category: Healthcare, Cameras, and Interactivity are just a few of the terms that come to mind when thinking about Computers processing the data collected by sensors. For example, in conjunction with the healthcare system, the heart monitor sensor regulates the cardiac rate. The application layer is a component of the IoMT network system that includes closed-circuit devices, wearables, and so on [41].

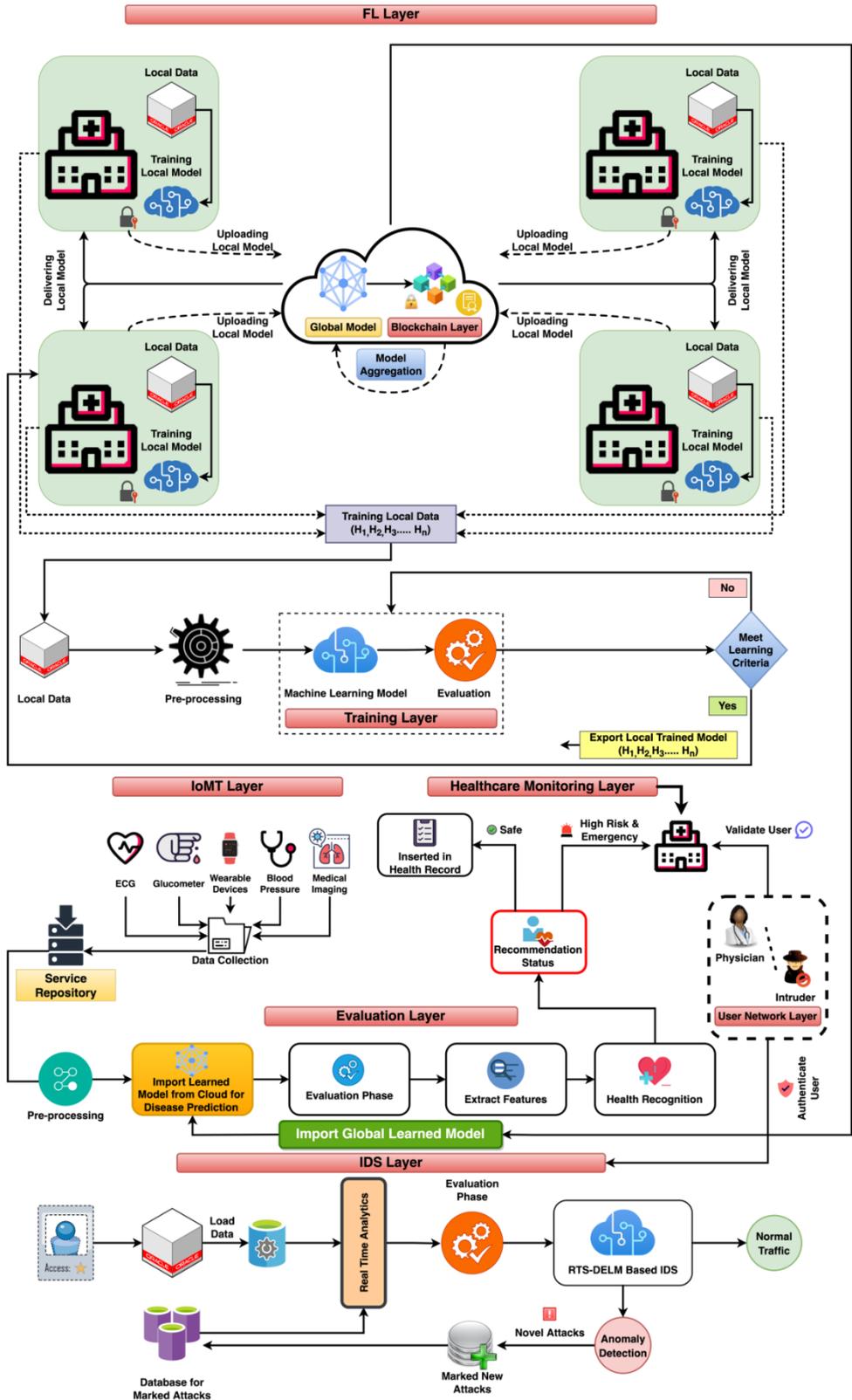

**Figure 3:** A Proposed Federated Learning-based Healthcare Monitoring System for Healthcare 5.0

## 3.5. Intrusion Detection System (IDS) Module Implementation

A comprehensive intrusion detection system is necessary to fully assess the healthcare system. Any type of data can be analyzed using the machine learning method termed RTS-DELM [42].

| No | Features | Form of value | No | Features | Form of value |
|---|---|---|---|---|---|
| 1 | Duration | Integer | 22 | is_guest_login | Integer |
| 2 | protocol_type | Nominal | 23 | count | Integer |
| 3 | Service | Nominal | 24 | srv_count | Integer |
| 4 | Flag | Nominal | 25 | serror_rate | Float |
| 5 | src_bytes | Integer | 26 | srv_serror_rate | Float |
| 6 | dst_bytes | Integer | 27 | rerror_rate | Float |
| 7 | land | Integer | 28 | srv_rerror_rate | Float |
| 8 | wrong_fragment | Integer | 29 | same_srv_rate | Float |
| 9 | urgent | Integer | 30 | diff_srv_rate | Float |
| 10 | hot | Integer | 31 | srv_diff_host_rate | Float |
| 11 | num_failed_logins | Integer | 32 | dst_host_count | Float |
| 12 | root_shell | Integer | 33 | dst_host_srv_count | Float |
| 13 | num_compromised | Integer | 34 | dst_host_same_srv_rate | Float |
| 14 | roots_hell | Integer | 35 | dst_host_diff_srv_rate | Float |
| 15 | su_attempted | Integer | 36 | dst_host_same_src_port_rate | Float |
| 16 | num_root | Integer | 37 | dst_host_srv_diff_port_rate | Float |
| 17 | num_file_creations | Integer | 38 | ddst_host_serror_rate | Float |
| 18 | num_shells | Integer | 39 | dst-_host_srv_serror_rate | Float |
| 19 | num_access_files | Integer | 40 | dst_host_rerror_rate | Float |
| 20 | num_outbound_cmds | Integer | 41 | dst_host_srv_rerror_rate | Float |
| 21 | Is_host_login | Integer | | | |

**Figure 4:** NSL-KDD Dataset Structure

| No | Features | No | Features | No | Features |
|---|---|---|---|---|---|
| 1 | MDVP:Fo(Hz) | 9 | MDVP:Shimmer | 17 | RPDE |
| 2 | MDVP:Fhi(Hz) | 10 | MDVP:Shimmer(dB) | 18 | DFA |
| 3 | MDVP:Flo(Hz) | 11 | Shimmer: APQ3 | 19 | spreadi |
| 4 | MDVP:Jitter(%) | 12 | Shimmer: APQ5 | 20 | spread2 |
| 5 | MDVP:Jitter(Abs) | 13 | MDVP:APO | 21 | D2 |
| 6 | MDVP:RAP | 14 | Shimmer:DDA | 22 | PPE |
| 7 | MDVP:PPO | 15 | NHR | 23 | Class Label |
| 8 | Jitter: DDP | 16 | HNR | | |

**Figure 5:** Parkinson's Disease Dataset Structure

The autonomous dataflow architecture used by this machine learning software allows it to track the information flow and identify trends of infiltration and assault. It's crucial to develop robust and adaptable algorithms to manage the continuously evolving smart blockchain-based applications. The suggested strategy provides a safe smart healthcare network design that would solve both present issues with centralized security of healthcare networks and potential future threats. In the work described here, a RTS-DELM technique will be employed to create a smarter, safer healthcare system employing sensors driven by the IoMT that are more effective as shown in                                    Figure 3. The main contributions of this study are a thorough analysis of scientific advancements feasible to blockchain-based healthcare 5.0 systems powered by RTS-DELM, a fresh perspective on various employments (like data exchange between healthcare systems), and assistance from the most recent stages of technological development [43].

## 3.6. Dataset Description

For disease prediction analysis, we employed the publicly accessible Parkinson's disease dataset [44], and for intrusion detection, we employed the publicly available NSL-KDD dataset [45]. **Error! Reference source**

**not found.** lists all protection protocols falling under their identity group. The NSL-KDD dataset is a refined edition of KDD 99 that includes several enhancements related to the original KDD 99 data collection. The NSL-KDD data collection contains 41 functions for each record. **Error! Reference source not found.** offers a proper description of the functions. This study includes 195 prolonged vowel phonations from 31 individuals, 23 of whom were diagnosed with Parkinson's disease. The primary objective of data processing is to distinguish between healthy individuals and those with Parkinson's disease, based on the "status" attribute, which is set to non-PD for healthy individuals and PD for those with Parkinson's disease; this is a two-decision classification issue. **Error! Reference source not found.** depicts the attributes of the data set.

*3.7. Working of the Proposed Model*

RTS-DELM computational technology may be used to make federated learning-based systems more intelligent. When using the RTS-DELM distributed blockchain technology, it is possible to increase the level of data secrecy. Additionally, by transferring new information and accelerating comprehension, RTS-DELM may be utilized to enhance understanding [30]. It provides the platform and network architecture for the development of a decentralized blockchain application [43]. The deployment architecture of the advanced system RTS-DELM is examined in this paper. Utilizing sensors, mobile devices, and IoMT systems as sources of information are the right way to employ this technology to gather intelligence. These strategies produce knowledge that is employed in smart applications. Despite this, real-time data analysis uses the RTS-DELM approach to assess and make predictions [42].

The creation of data for study reduces data mistakes such as repetition, missing information factors, malfunctions, and interference. When a little fraction of a data collection is all that is needed, the RTS-DELM method performs effectively. The architecture is capable of supporting a broad variety of applications in several fields, including fraud detection and/or prevention. As shown in Figure 3, the proposed RTS-DELM system employs a large number of hidden layers, hidden neurons, and various activating mechanisms to optimize the healthcare monitoring system. Data capture, preparation, and assessment are the three separate phases that make up the suggested approach for breaking down data analysis. Prediction and performance make up the two sub-layers that make up the evaluation layer. Accurate data is gathered from sensors and actuators for analysis. The collecting layer uses the data that is provided as raw data. To eliminate inconsistencies in the preprocessing layer, a thorough approach for data cleaning and preparation is used. The foremost objectives of this approach are:

- To implement the RTS-DELM algorithm to build an optimal approach for healthcare monitoring at the client side.
- In this study, an intelligent algorithm is proposed to identify and monitor healthcare in patients and identify the severity of the patients.
- To implement an Intrusion Detection System (IDS) to determine the flow of data to detect intrusions and attack patterns
- To implement and improve IoMT in existing applications to sustain the medical standards. Furthermore, to increase the model's efficiency, a federated learning approach is implemented as shown in table 2 & table 3. Further evaluation is carried out using disease datasets and the freshly trained layer is applied to the pre-trained framework to increase the efficacy of the system.
- To preserve records of the local data to preserve the patient's privacy.
- To test the utility of a trained model by assessing it on a real-time dataset. with federated learning, the implementation of the prediction system is evaluated. In addition, model output is evaluated with other models of machine learning.
- To prevent humans from getting chronic diseases by the spreading of the disease.
- To estimate the performance of the proposed methodology by assessing different datasets.
- Furthermore, test and evaluate the datasets on different machine learning algorithms to justify the proposed approach.

The study will provide a thorough analysis of technical advancements relevant to RTS-DELM-enabled federated learning-based systems to provide a fresh perspective on various implementations (such as healthcare data exchange). The main aim of this study is to build an intelligent algorithm proposed for healthcare monitoring among patients. The working of the proposed model is described as follows;

- Hospitals identify and assign training tasks to a local model, which is subsequently uploaded to a centralized server, where it is disseminated to all IoMT devices as a global model.
- Additionally, the training phase comprises three layers: the sensing layer, the preprocessing layer, and the application layer.
- Medical data obtained by IoMT devices may contain missing or erroneous data.
- To reduce noisy data, the preprocessing layer uses moving average and normalization to address missing values.
- Clinical data is transmitted to the Application layer following preparation. Additionally, the Application Layer is separated into two sections: Prediction Layer and Performance Layer.
- Following the prediction layer, the results of the forecast layer are transmitted to the performance layer, which uses the accuracy and miss rate of the prediction layer to determine whether or not the learning requirements are met.
- In the case of 'No' the model will be retrained but in the case of 'Yes', the locally trained model is exported to the cloud as a global model.
- Medical organizations receive a copy of a global model and train it on local data. Following that, each institution upgrades its model on the cloud without exchanging datasets directly. This assures the continuity of patient privacy between medical organizations.
- The cloud aggregates all updated parameters provided by participating institutions to create a new global model, which is subsequently disseminated to all participating organizations.
- As opposed to outcomes produced from a single information source, the data fusion technique can give results that are more dependable and stable.
- Then the data requester launches a data sharing request to the centralized server. Upon receiving the request, the centralized server verifies the access. After authorizing access, the nodes train a global data model jointly by federated learning. Once the model is trained, the data requester gets the corresponding sharing results.
- The input layer parameters will be detected during the validation phase and forwarded to the evaluation phase for healthcare monitoring.

**Table 2:** Proposed RTS-DELM Entangled with Federated Learning Pseudo code (Server-side)

| [Server Side] | |
|---|---|
| Sr No. | Steps |
| 1 | Start |
| 2 | Initialize $w_{G,fml}^k$ & $v_{G,fml}^k$ <br> **Where** $w_{G,fml}^k$ & $v_{G,fml}^k$ represents the weight between input and $y$ hidden layers at server side and the weight between $y$ hidden layers and $y+1$ hidden layers neurons at server side respectively. |
| 3 | **for** each cycle $k$ from $I$ to $K$ **do** |
| 4 | $S_k \leftarrow$ (Random set of clients from $\eta$ ) |
| a) | **for** each client $l \in S_k$ parallelly **do** <br><br> $[w_{k+1}^n, v_{k+1}^n] \leftarrow$ **Client Training** $(n, w_k, v_k)$ <br><br> **end for** |
| 5 | $w_{G,fml}^k = \frac{1}{\Sigma_{n \varepsilon \eta}} \Sigma_{n=1}^N \frac{S_n}{S} w_{n+1}^k$ (Avg Aggregation) |

| 6. | $v_{G,fml}^k = \frac{1}{\sum_{n\in\eta}} \Sigma_{n=1}^N \frac{S_n}{S} v_{n+1}^k$ |
|---|---|
| 7. | end for |
| 8. | Stop |

Put the optimum weights values of $w_{G,fml}^k$ & $v_{G,fml}^k$ in Equations 1 & 2, and predict the values for lung cancer found or not.

Table 3: Proposed Smart FML- RTS-DELM Pseudo code (Client-side)

| [Client Training ($L, w, v$) ] | |
|---|---|
| Sr No. | Steps |
| 1 | Start |
| 2 | Split local data to mini batches of size S |
| 3 | Initialization of both layer weights ($\omega_{ij}$ & $v_{jk}$), Error (E)= 0 and the number of epochs £ = 0 |
| 4 | For each training pattern p |
|  | b) do the feedforward phase to |
|  | i) calculate ɸ$_j$ using eq (1) |
|  | ii) calculate ɸ$_k$ using eq (2) |
|  | c) Calculate error signals for the output and the hidden layer |
|  | d) Then equalize the weights $v_{jk}$ and $\omega_{ij}$ (backpropagation of errors) using eq (8) & eq (9). |
| 5 | £ = £ + 1 |
| 6. | Test stopping criteria: if no stopping criterion is satisfied, go to step 4. |
| 7. | Return optimum local trained model weights $\omega_{ij}$ and $v_{jk}$ to Server |
| 8. | Stop |

Each client Proposed RTS-DELM has used input layer, six hidden layers and an output layer as exhibited in table 3. The backpropagation algorithm has several phases, including initialization of weight, feedforward, backpropagation of error, and updating of weight and bias, as indicated in the table above. Each neuron in the hidden layer is equipped with a Sigmoid activation function. The proposed system based on RTS-DELM may be stated [40];

$$ɸ_{jl} = \frac{1}{1+e^{-(b_y+\Sigma_{i=1}^m(\omega_{ij}*r_i))}} \quad where\ j = 1,2,3 \dots n \tag{1}$$

**Where** $r_i$ input data, $b_1$ is bias, $m$ represents total number of input neurons and $j$ represents total number of hidden layer neurons.

Output layer activation function is given below [40];

$$ɸ_k = \frac{1}{1+e^{-(b_2+\Sigma_{j=1}^n(v_{jk}y*ɸ_j))}} \quad where\ k = 1,2,3 \dots r \tag{2}$$

**Where** $y$ represents hidden layers [41];

$$E = \frac{1}{2}\Sigma_k(\tau_k - ɸ_{ky=6})^2 \tag{3}$$

Above equation $E$ represents backpropagation error where, $\tau_k$ & ɸ$_k$ symbolize the anticipated output and projected output.

In equation (4), the weight of the output changes at a constant pace [41], the layer is composed as;

$$\Delta W \propto - \frac{\partial E}{\partial W}$$

$$\Delta v_{j,k^y} = -\epsilon \frac{\partial E}{\partial v_{j,k^y}} \tag{4}$$

After employing the Chain rule technique above eq can be composed as [42];

$$\Delta v_{j,k^y} = -\epsilon \frac{\partial E}{\partial \varphi_{k^y}} \times \frac{\partial \varphi_{k^y}}{\partial v_{j,k^y}} \tag{5}$$

After substituting the values in equation (5), the value of weight changed can be obtained as shown in equation (6) [40] [42].

$$\Delta v_{j,k^y} = \epsilon(\tau_k - \varphi_{k^y}) \times \varphi_{k^y}(1 - \varphi_{k^y}) \times (\varphi_j)$$

$$\Delta v_{j,k^y} = \epsilon \xi_k \varphi_j \tag{6}$$

Where,

$$\xi_k = (\tau_k - \varphi_{k^y}) \times \varphi_{k^y}(1 - \varphi_{k^y})$$

Utilize the chain rule to maintain the weights between the input and hidden layers [41]

$$\Delta \omega_{i,j} \propto - \left[\sum_k \frac{\partial E}{\partial \varphi_{k^y}} \times \frac{\partial \varphi_{k^y}}{\partial \varphi_j}\right] \times \frac{\partial \varphi_j}{\partial \omega_{i,j}}$$

$$\Delta \omega_{i,j} = -\epsilon \left[\sum_k \frac{\partial E}{\partial \varphi_{k^y}} \times \frac{\partial \varphi_{k^y}}{\partial \varphi_j}\right] \times \frac{\partial \varphi_j}{\partial \omega_{i,j}}$$

In the above eq, $\epsilon$ symbolizes the constant [41],

$$\Delta \omega_{i,j} = \epsilon \left[\sum_k (\tau_k - \varphi_{k^y}) \times \varphi_{k^y}(1 - \varphi_{k^y}) \times (v_{j,k})\right] \times \varphi_{k^y}(1 - \varphi_{k^y}) \times \alpha_i$$

$$\Delta \omega_{i,j} = \epsilon \left[\sum_k (\tau_k - \varphi_{k^y}) \times \varphi_{k^y}(1 - \varphi_{k^y}) \times (v_{j,k})\right] \times \varphi_j(1 - \varphi_j) \times \alpha_i$$

$$\Delta \omega_{i,j} = \epsilon \left[\sum_k \xi_k (v_{j,k^y})\right] \times \varphi_j(1 - \varphi_j) \times \alpha_i$$

After simplifying the preceding equation, it can be expressed as [42]

$$\Delta \omega_{i,j} = \epsilon \xi_j \alpha_i \tag{7}$$

Where,

$$\xi_j = \left[\sum_k \xi_k (v_{j,k^y})\right] \times \varphi_j(1 - \varphi_j)$$

$$v_{i,k^y}^+ = v_{j,k^y} + \lambda \Delta v_{j,k^y} \tag{8}$$

**Where $\lambda$ represents the learning rate**

The above equation is used for updating the weights between output and hidden layers and the below equation is used for updating the weights between input and hidden layers [40][41][42].

$$w_{i,k}^+ = w_{i,j} + \lambda \Delta w_{i,j} \tag{9}$$

## 4. Simulation Results

In this research, the healthcare 5.0 system entangled with federated learning approach was applied to the Parkinson's disease dataset [44] for Parkinson's disease prediction and the NSL-KDD dataset [45] for detecting any intrusion activity in a system. The datasets are randomly divided into 70% of training and 30% of data is used for validation and testing. The data is investigated in diagnosing Parkinson's disease and detecting intrusion activity in a system. To estimate the efficacy of the proposed method, numerous numerical metrics like Miss rate, accuracy, specificity, sensitivity, True Positive Rate (TRP), True Negative Rate (TNR), Positive Prediction Value (PPV) & Negative Prediction Value (NPV) were employed listed as follows [40] [41] [42];

$$\text{Miss rate} = \frac{\sum_{b=0}^{2}(Q_b/S_{z \neq b})}{\sum_{b=0}^{2}(T_b)}, \quad Where\ z = 0,1 \tag{10}$$

$$\text{Accuracy} = \frac{\sum_{b=0}^{2}(Q_b/S_b)}{\sum_{b=0}^{2}(Q_b)} \tag{11}$$

$$\text{Sepecificity} = \frac{Q_0/S_0}{(Q_0/S_0 + Q_0/S_1)} \tag{12}$$

$$\text{Sensitivity} = \frac{Q_1/V_1}{(Q_1/S_0 + Q_1/S_1)} \tag{13}$$

$$TPR = \frac{TP}{(TP+FN)} \tag{14}$$

$$TNR = \frac{TN}{(TN+FP)} \tag{15}$$

$$PPV = \frac{TP}{(TP+FP)} \tag{16}$$

$$NPV = \frac{TN}{(FN+TN)} \tag{17}$$

In Eq. (10) and (11), $Q$ signifies the predictive output and $S$ symbolizes the actual output. $Q_0$ and $S_0$ represents that there is no Parkinson's disease and there is no intrusion activity in predictive output and actual output correspondingly. $Q_1$ and $S_1$ signifies there is a Parkinson's disease and intrusion activity in the predictive result and actual result individually. $Q\_b / S\_b$ denotes predictive and actual results are parallel. Correspondingly, $Q\_b / S\_(z \neq b)$ epitomizes error, whereby outcomes, both predictive and actual, are altered.

### 4.1. Intrusion Detection System (IDS)

Table 4 exhibited the suggested RTS-DELM-based secure healthcare 5.0 system for the prediction of intrusion in the system during the training level. Through the training phase, a total of 400 records are applied at each client side ($H_1$, $H_2$, $H_3$, $H_4$), which are technically separated into normal and attack records, respectively. It is seen that the forecasting system successfully achieve maximum accuracy for forecast the intrusion in a system at each client side. Table 4 shows the different statistical measures during training level at each client side. As shown in Table 4, proposed system is effective in terms of accuracy to determine the intrusion at each client side. In addition, some other statistical measures are calculated during training level. As we can see that in Table 4, $H_1$ client gives the 93.75% accuracy, 98.25% sensitivity, 82.61% specificity, 95% NPV, 17.39% FPR, 6.67% FDR and 1.75% FNR. While $H_2$ client gives the 94.72% accuracy, 98.95% sensitivity, 84.07% specificity, 96.94% NPV, 15.93% FPR, 6% FDR and 1.05% FNR. While $H_3$ client gives the 97.75% accuracy, 98.99% sensitivity, 94.17% specificity, 97% NPV, 5.83% FPR, 2% FDR and 1.01% FNR. While $H_4$ client gives the 95.72% accuracy, 99.30% sensitivity, 86.36% specificity, 97.94% NPV, 13.64% FPR, 5% FDR and 0.7% FNR.

**Table 4:** Performance Evaluation of Proposed RTS-DELM-Based Secure Healthcare 5.0 System During Training for the Estimation of Intrusion Detection in a System Using Different Statistical Measurements at Client Side

| Client | Accuracy | Sensitivity | Specificity | Negative Predictive Value (NPV) | False Positive Rate (FPR) | False Discovery Rate (FDR) | False Negative Rate (FNR) |
|---|---|---|---|---|---|---|---|
| $H_1$ | 0.9375 | 0.9825 | 0.8261 | 0.9500 | 0.1739 | 0.0667 | 0.0175 |
| $H_2$ | 0.9472 | 0.9895 | 0.8407 | 0.9694 | 0.1593 | 0.0600 | 0.0105 |
| $H_3$ | 0.9775 | 0.9899 | 0.9417 | 0.9700 | 0.0583 | 0.0200 | 0.0101 |
| $H_4$ | 0.9572 | 0.9930 | 0.8636 | 0.9794 | 0.1364 | 0.0500 | 0.0070 |

Table 5 exhibited the suggested RTS-DELM-based secure healthcare 5.0 system for the prediction of intrusion in the system during the validation level. Through the validation phase, a total of 200 records are applied at each client side ($H_1$, $H_2$, $H_3$, $H_4$), which are technically separated into normal and attack records, respectively. It is seen that the detection system successfully achieves maximum accuracy for forecast the intrusion in a system at each client side. Table 5 shows the different statistical measures during validation level at each client side. As shown in Table 5, proposed system is effective in terms of accuracy to determine the intrusion at each client side. In addition, some other statistical measures are calculated also during validation level to justify the effectiveness of the proposed system. As we can see that in Table 5, $H_1$ client gives the 95% accuracy, 97.30% sensitivity, 88.64% specificity, 92% negative predictive value, 11.54% false positive rate, 4% false discovery rate and 2.7% false negative rate. While $H_2$ client gives the 93.50% accuracy, 95.75% sensitivity, 86.27% specificity, 88% negative predictive value, 13.73% false positive rate, 4.67% false discovery rate and 4.03% false negative rate. While $H_3$ client gives the 96.50% accuracy, 98.64% sensitivity, 90.57% specificity, 96% negative predictive value, 9.43% false positive rate, 3.33% false discovery rate and 1.36% false negative rate. While $H_4$ client gives the 94.50% accuracy, 96.64% sensitivity, 88.24% specificity, 90% negative predictive value, 11.76% false positive rate, % false discovery rate and 3.36% false negative rate.

**Table 5:** Performance Evaluation of Proposed RTS-DELM-Based Secure Healthcare 5.0 System During Validation for the Estimation of Intrusion Detection in a System Using Different Statistical Measurements at Client Side

| Client | Accuracy | Sensitivity | Specificity | Negative Predictive Value | False Positive Rate | False discovery Rate | False Negative Rate |
|---|---|---|---|---|---|---|---|
| $H_1$ | 0.9500 | 0.9730 | 0.8846 | 0.9200 | 0.1154 | 0.0400 | 0.0270 |
| $H_2$ | 0.9350 | 0.9597 | 0.8627 | 0.8800 | 0.1373 | 0.0467 | 0.0403 |
| H₃ | 0.9650 | 0.9864 | 0.9057 | 0.9600 | 0.0943 | 0.0333 | 0.0136 |
| H₄ | 0.9450 | 0.9664 | 0.8824 | 0.9000 | 0.1176 | 0.0400 | 0.0336 |

*4.2. Disease Prediction*

Table 6 exhibited the suggested healthcare 5.0 system entangled with federated learning approach for the prediction of Parkinson's disease in the patients during the validation level. Each client side ($H_1$, $H_2$, $H_3$, $H_4$) at training level trained their model on their local data and exported the learned model to cloud to form a global learned model. Furthermore, the global learned model imported from blockchain centered centralized server to validate the proposed system after verifying access through IDS. Through the validation phase, a total of 200 records are applied at server side, which are technically separated into negative and positive samples, respectively. It is seen that the forecasting system successfully achieve maximum accuracy for predict the disease in a patient at server side. Table 6 shows the different statistical measures during validation level at each client side and server side utilizing FL approach to justify the proposed healthcare 5.0 system entangled with federated learning approach in terms of accuracy. As presented in Table 6, proposed healthcare 5.0 system entangled with federated learning system is effective in terms of accuracy to predict the disease. In addition, some other statistical measures are calculated during validation level. As we can see that in Table 6, $H_1$ client gives the 92.50% accuracy, 96.97% sensitivity, 71.43% specificity, 83.33% negative predictive value, 28.57% false positive rate, 5.88% false discovery rate and 3.03% false negative rate during validation level. While $H_2$ client gives the 93% accuracy, 96.43% sensitivity, 75% specificity, 80% negative predictive value, 25% false positive rate, 4.71% false discovery rate and 3.57% false negative rate during validation level. While $H_3$ client gives the 95.50% accuracy, 97.63% sensitivity, 83.87% specificity, 86.67% negative predictive value, 16.63% false positive rate, 2.94% false discovery rate and 2.37% false negative rate during validation level. While $H_4$ client gives the 94.50% accuracy, 96.49% sensitivity, 82.76% specificity, 80% negative predictive value, 17.24% false positive rate, 2.94% false discovery rate and 3.51% false negative rate during validation level. Finally, at server side the proposed healthcare 5.0 system entangled with federated learning approach achieves the maximum accuracy as compare to each client. The proposed FL approach gives the 97% accuracy, 98.24% sensitivity, 90% specificity, 90% negative predictive value, 10% false positive rate, 1.76% false discovery rate and 1.76% false negative rate during validation level.

**Table 6:** Performance Evaluation of Proposed Healthcare 5.0 System Entangled with Federated Learning Approach During Validation for the Prediction of Disease in a Patient Using Different Statistical Measurements at Server Side

| Client | Accuracy | Sensitivity | Specificity | Negative Predictive Value | False Positive Rate | False discovery Rate | False Negative Rate |
|---|---|---|---|---|---|---|---|
| $H_1$ | 0.9250 | 0.9697 | 0.7143 | 0.8333 | 0.2857 | 0.0588 | 0.0303 |
| $H_2$ | 0.9300 | 0.9643 | 0.7500 | 0.8000 | 0.2500 | 0.0471 | 0.0357 |
| $H_3$ | 0.9550 | 0.9763 | 0.8387 | 0.8667 | 0.1613 | 0.0294 | 0.0237 |
| $H_4$ | 0.9450 | 0.9649 | 0.8276 | 0.8000 | 0.1724 | 0.0294 | 0.0351 |
| **Proposed Approach based on FL (Server Side)** | **0.9700** | **0.9824** | **0.9000** | **0.9000** | **0.1000** | **0.0176** | **0.0176** |

## 5. Discussion

Table 7 shows the comparison of the proposed model with previously published research. Sheibani et al. [46] achieves 90.6% accuracy using 10-fold cross-validation. Tracy et al. [47] achieves 90.1% accuracy using L2-regularized logistic regression, random forest. Sztahó et al. [48] achieves 89.3% accuracy using KNN, SVM-linear, SVM-RBF, ANN and DNN. Yaman et al. [49] achieves 91.25% accuracy using KNN, SVM with 10-fold cross-validation. Kuresan et al. [50] achieves 95.16% accuracy using HMM, SVM. Chang et al. [38] achieves 84.5% accuracy using CNN. While proposed healthcare 5.0 system entangled with federated learning approach model achieves 97% accuracy. As shown in Table 7, the proposed approach outclasses other approaches in terms of accuracy.

**Table 7:** Performance evaluation of proposed RTS-DELM Based Architecture for the Estimation of Parkinson's Disease with previously published research

|  | **Machine Learning Methods** | **Outcome** |
|---|---|---|
| Sheibani et al. [46] | Ensemble learning with 10-fold cross-validation | 90.6% |
| Tracy et al. [47] | L2-regularized logistic regression, random forest | 90.1% |
| Sztahó et al. [48] | KNN, SVM-linear, SVM-RBF, ANN, DNN | 89.3% |
| Yaman et al. [49] | KNN, SVM with 10-fold cross-validation | 91.25% |
| Kuresan et al. [50] | HMM, SVM | 95.16% |
| Chang et al. [38] | Convolutional Neural Network (CNN) | 84.5% |
| **Proposed Model** | **Healthcare 5.0 System Entangled with Federated Learning Approach** | **97%** |

## 6. Conclusion

In recent years, IoMT-based intelligent healthcare has been widely implemented to maximize the utilization of medical data to improve the accuracy of disease forecasting and healthcare therapy. With the rising amount and diversity of medical data, there is an urgent need for efficient data mining techniques to assess this information in order to facilitate disease identification, provide medical treatments, and improve patient care. Nonetheless, it confronts several difficulties, including patient privacy breaches and a variety of adversary threats to data transmission. In addition, the development of artificial intelligence and worldwide epidemics have hastened the adoption of intelligent healthcare while presenting problems around information security, hostile cyberattacks, and service quality. A secure healthcare 5.0 system based on blockchain technology entangled with federated learning techniques is being explored to increase predictive performance. Numerous analytical frameworks have been employed to determine the viability of this particular argument. The proposed RTS-DELM method is extremely effective. The proposed technique has a rate of accuracy of 93.22 percent for disease prediction and 96.18 percent for intrusion detection. Furthermore, it is acknowledged that developing a basic method is less expensive and faster. We are pleased with the preliminary findings and intend to expand our research by evaluating additional sets of data in the future. The proposed system's computational complexity is limited by the growing number of hidden layers. Future research will aim to identify and quantify the factors with greater precision. To optimize the performance of various configurations, the learning system will be retrained on a more frequent basis.

## References

[1]  C. Wilson, T. Hargreaves, R. Hauxwell-Baldwin, Benefits and risks of smart home technologies, Energy Policy. 103 (2017) 72-83. https://doi.org/10.1016/j.enpol.2016.12.047.

[2]  B.L. Risteska Stojkoska, K. V. Trivodaliev, A review of Internet of Things for smart home: Challenges and solutions, J. Clean. Prod. 140 (2017) 1454-1464. https://doi.org/10.1016/j.jclepro.2016.10.006.

[3]  F. Folianto, Y.S. Low, W.L. Yeow, Smartbin: Smart waste management system, in: 2015 IEEE 10th Int. Conf. Intell. Sensors, Sens. Networks Inf. Process. ISSNIP 2015, 2015. https://doi.org/10.1109/ISSNIP.2015.7106974.

[4]  J. ho Park, M.M. Salim, J.H. Jo, J.C.S. Sicato, S. Rathore, J.H. Park, CIoT-Net: a scalable cognitive IoT based smart city network architecture, Human-Centric Comput. Inf. Sci. 9 (2019) 1-20. https://doi.org/10.1186/s13673-019-0190-9.

[5]  M.R. Alam, M. St-Hilaire, T. Kunz, Peer-to-peer energy trading among smart homes, Appl. Energy. 238 (2019) 1434-1443. https://doi.org/10.1016/j.apenergy.2019.01.091.

[6]  Y. Mittal, P. Toshniwal, S. Sharma, D. Singhal, R. Gupta, V.K. Mittal, A voice-controlled multi-functional Smart Home Automation System, in: 12th IEEE Int. Conf. Electron. Energy, Environ. Commun. Comput. Control (E3-C3), INDICON 2015, 2016. https://doi.org/10.1109/INDICON.2015.7443538.

[7]  P. Wang, F. Ye, X. Chen, A Smart Home Gateway Platform for Data Collection and Awareness, IEEE Commun. Mag. 56 (2018) 87-93. https://doi.org/10.1109/MCOM.2018.1701217.

[8]  J. Shen, C. Wang, T. Li, X. Chen, X. Huang, Z.H. Zhan, Secure data uploading scheme for a smart home system, Inf. Sci. (Ny). 453 (2018) 186-197. https://doi.org/10.1016/j.ins.2018.04.048.

[9]  N. Komninos, E. Philippou, A. Pitsillides, Survey in smart grid and smart home security: Issues, challenges and countermeasures, IEEE Commun. Surv. Tutorials. 16 (2014) 1933-1954. https://doi.org/10.1109/COMST.2014.2320093.

[10] S. Abbas, M.A. Khan, L.E. Falcon-Morales, A. Rehman, Y. Saeed, M. Zareei, A. Zeb, E.M. Mohamed, Modeling, Simulation and Optimization of Power Plant Energy Sustainability for IoT Enabled Smart Cities Empowered with Deep Extreme Learning Machine, IEEE Access. 8 (2020) 39982-39997. https://doi.org/10.1109/ACCESS.2020.2976452.

[11] D. Nasonov, A.A. Visheratin, A. Boukhanovsky, Blockchain-based transaction integrity in distributed big data marketplace, in: Lect. Notes Comput. Sci. (Including Subser. Lect. Notes Artif. Intell. Lect. Notes Bioinformatics), 2018. https://doi.org/10.1007/978-3-319-93698-7_43.

[12] R.A. Michelin, A. Dorri, M. Steger, R.C. Lunardi, S.S. Kanhere, R. Jurdak, A.F. Zorzo, SpeedyChain: A framework for decoupling data from blockchain for smart cities, in: ACM Int. Conf. Proceeding Ser., 2018. https://doi.org/10.1145/3286978.3287019.

[13] B. Xiong, K. Yang, J. Zhao, K. Li, Robust dynamic network traffic partitioning against malicious attacks, J. Netw. Comput. Appl. 87 (2017) 20-31. https://doi.org/10.1016/j.jnca.2016.04.013.

[14] C. Yin, J. Xi, R. Sun, J. Wang, Location privacy protection based on differential privacy strategy for big data in industrial internet of things, IEEE Trans. Ind. Informatics. 14 (2018) 3628-3636. https://doi.org/10.1109/TII.2017.2773646.

[15] Z. Zheng, S. Xie, H.N. Dai, X. Chen, H. Wang, Blockchain challenges and opportunities: A survey, Int. J. Web Grid Serv. 14 (2018) 352-375. https://doi.org/10.1504/IJWGS.2018.095647.

[16] M. Rahouti, K. Xiong, N. Ghani, Bitcoin Concepts, Threats, and Machine-Learning Security Solutions, IEEE Access. 6 (2018) 67189-67205. https://doi.org/10.1109/ACCESS.2018.2874539.

[17] B. Mohanta, P. Das, S. Patnaik, Healthcare 5.0: A paradigm shift in digital healthcare system using artificial intelligence, IOT and 5G communication, in: Proc. - 2019 Int. Conf. Appl. Mach. Learn. ICAML 2019, 2019. https://doi.org/10.1109/ICAML48257.2019.00044.

[18] E. Mbunge, B. Muchemwa, S. Jiyane, J. Batani, Sensors and healthcare 5.0: transformative shift in virtual care through emerging digital health technologies, Glob. Heal. J. 5 (2021) 169-177. https://doi.org/10.1016/j.glohj.2021.11.008.

[19] M. Bhavin, S. Tanwar, N. Sharma, S. Tyagi, N. Kumar, Blockchain and quantum blind signature-based hybrid scheme for healthcare 5.0 applications, J. Inf. Secur. Appl. 56 (2021) p.102673. https://doi.org/10.1016/j.jisa.2020.102673.

[20] S. Aggarwal, R. Chaudhary, G.S. Aujla, N. Kumar, K.K.R. Choo, A.Y. Zomaya, Blockchain for smart communities: Applications, challenges and opportunities, J. Netw. Comput. Appl. 144 (2019) 13-48. https://doi.org/10.1016/j.jnca.2019.06.018.

[21] M. Andoni, V. Robu, D. Flynn, S. Abram, D. Geach, D. Jenkins, P. McCallum, A. Peacock, Blockchain technology in the energy sector: A systematic review of challenges and opportunities, Renew. Sustain. Energy Rev. 100 (2019) 143-174. https://doi.org/10.1016/j.rser.2018.10.014.

[22] G. Li, M. Dong, L.T. Yang, K. Ota, J. Wu, J. Li, Preserving Edge Knowledge Sharing among IoT Services: A Blockchain-Based Approach, IEEE Trans. Emerg. Top. Comput. Intell. 4 (2020) 653-665. https://doi.org/10.1109/TETCI.2019.2952587.

[23] Z. Zhou, B. Wang, M. Dong, K. Ota, Secure and Efficient Vehicle-to-Grid Energy Trading in Cyber Physical Systems: Integration of Blockchain and Edge Computing, IEEE Trans. Syst. Man, Cybern. Syst. 50 (2020) 43-57. https://doi.org/10.1109/TSMC.2019.2896323.

[24] X. Du, B. Chen, M. Ma, Y. Zhang, Research on the Application of Blockchain in Smart Healthcare: Constructing a Hierarchical Framework, J. Healthc. Eng. 2021 (2021). https://doi.org/10.1155/2021/6698122.

[25] B. Ihnaini, M.A. Khan, T.A. Khan, S. Abbas, M.S. Daoud, M. Ahmad, M.A. Khan, A Smart Healthcare Recommendation System for Multidisciplinary Diabetes Patients with Data Fusion Based on Deep Ensemble Learning, Comput. Intell. Neurosci. 2021 (2021). https://doi.org/10.1155/2021/4243700.

[26] M.A. Khan, Challenges Facing the Application of IoT in Medicine and Healthcare, Int. J. Comput. Inf. Manuf. 1 (2021). https://doi.org/10.54489/ijcim.v1i1.32.

[27] M.F. Khan, T.M. Ghazal, R.A. Said, A. Fatima, S. Abbas, M.A. Khan, G.F. Issa, M. Ahmad, M.A. Khan, An iomt-enabled smart healthcare model to monitor elderly people using machine learning technique, Comput. Intell. Neurosci. 2021 (2021). https://doi.org/10.1155/2021/2487759.

[28] J. Xu, B.S. Glicksberg, C. Su, P. Walker, J. Bian, F. Wang, Federated Learning for Healthcare Informatics, J. Healthc. Informatics Res. 5 (2021) 1-19. https://doi.org/10.1007/s41666-020-00082-4.

[29] Y. Li, B. Shan, B. Li, X. Liu, Y. Pu, Literature Review on the Applications of Machine Learning and Blockchain Technology in Smart Healthcare Industry: A Bibliometric Analysis, J. Healthc. Eng. 2021 (2021). https://doi.org/10.1155/2021/9739219.

[30] S.Y. Siddiqui, I. Naseer, M.A. Khan, M.F. Mushtaq, R.A. Naqvi, D. Hussain, A. Haider, Intelligent breast cancer prediction empowered with fusion and deep learning, Comput. Mater. Contin. 67 (2021). https://doi.org/10.32604/cmc.2021.013952.

[31] H. Medjahed, D. Istrate, J. Boudy, J.L. Baldinger, B. Dorizzi, A pervasive multi-sensor data fusion for smart home healthcare monitoring, in: IEEE Int. Conf. Fuzzy Syst., 2011. https://doi.org/10.1109/FUZZY.2011.6007636.

[32] W. Dai, T.S. Brisimi, W.G. Adams, T. Mela, V. Saligrama, I.C. Paschalidis, Prediction of hospitalization due to heart diseases by supervised learning methods, Int. J. Med. Inform. 84 (2015) 189-197. https://doi.org/10.1016/j.ijmedinf.2014.10.002.

[33] Y.J. Son, H.G. Kim, E.H. Kim, S. Choi, S.K. Lee, Application of support vector machine for prediction of medication adherence in heart failure patients, Healthc. Inform. Res. 16 (2010) 253-259. https://doi.org/10.4258/hir.2010.16.4.253.

[34] A. Tariq, L.A. Celi, J.M. Newsome, S. Purkayastha, N.K. Bhatia, H. Trivedi, J.W. Gichoya, I. Banerjee, Patient-specific COVID-19 resource utilization prediction using fusion AI model, Npj Digit. Med. 4 (2021) 1-9. https://doi.org/10.1038/s41746-021-00461-0.

[35] A. Sedik, A.M. Iliyasu, B.A. El-Rahiem, M.E. Abdel Samea, A. Abdel-Raheem, M. Hammad, J. Peng, F.E. Abd El-Samie, A.A. Abd El-Latif, Deploying machine and deep learning models for efficient data-augmented detection of COVID-19 infections, Viruses. 12 (2020). https://doi.org/10.3390/v12070769.

[36] Qayyum, A., Ahmad, K., Ahsan, M.A., Al-Fuqaha, A, Qadir, J, Collaborative federated learning for healthcare: Multi-modal covid-19 diagnosis at the edge. (2021) arXiv preprint arXiv:2101.07511.

[37] T.S. Brisimi, R. Chen, T. Mela, A. Olshevsky, I.C. Paschalidis, W. Shi, Federated learning of predictive models from federated Electronic Health Records, Int. J. Med. Inform. 112 (2018) 59-67. https://doi.org/10.1016/j.ijmedinf.2018.01.007.

[38] Y. Chang, C. Fang, W. Sun, A blockchain-based Federated Learning Method for Smart Healthcare, Computational Intelligence and Neuroscience. 2021 (2021) 1–12. doi:10.1155/2021/4376418.

[39] S. Squarepants, Bitcoin: A Peer-to-Peer Electronic Cash System, SSRN Electron. J. (2022). https://doi.org/10.2139/ssrn.3977007.

[40] A. Rehman, A. Athar, M.A. Khan, S. Abbas, A. Fatima, Atta-Ur-Rahman, A. Saeed, Modelling, simulation, and optimization of diabetes type II prediction using deep extreme learning machine, J. Ambient Intell. Smart Environ. 12 (2020) 125-138. https://doi.org/10.3233/AIS-200554.

[41] M.A. Khan, A. Rehman, K.M. Khan, M.A. Al Ghamdi, S.H. Almotiri, Enhance intrusion detection in computer networks based on deep extreme learning machine, Comput. Mater. Contin. 66 (2021). https://doi.org/10.32604/cmc.2020.013121.

[42] A. Haider, M.A. Khan, A. Rehman, M. Ur Rahman, H.S. Kim, A real-time sequential deep extreme learning machine cybersecurity intrusion detection system, Comput. Mater. Contin. 66 (2020). https://doi.org/10.32604/cmc.2020.013910.

[43] M.A. Khan, S. Abbas, A. Rehman, Y. Saeed, A. Zeb, M.I. Uddin, N. Nasser, A. Ali, A Machine Learning Approach for Blockchain-Based Smart Home Networks Security, IEEE Netw. 35 (2021) 223-229. https://doi.org/10.1109/MNET.011.2000514.

[44] Oxford, Parkinsons Data Set, UCI Learn. Repos. (2007).

[45] A.G. M. Tavallaee, E. Bagheri, W. Lu, Canadian Institute for Cybersecurity, UNB, NSL-KDD Dataset. (2018).

[46] R. Sheibani, E. Nikookar, S. Alavi, An ensemble method for diagnosis of Parkinson's disease based on voice measurements, J. Med. Signals Sens. 9 (2019). https://doi.org/10.4103/jmss.JMSS_57_18.

[47] J.M. Tracy, Y. Özkanca, D.C. Atkins, R. Hosseini Ghomi, Investigating voice as a biomarker: Deep phenotyping methods for early detection of Parkinson's disease, J. Biomed. Inform. 104 (2020). https://doi.org/10.1016/j.jbi.2019.103362.

[48] D. Sztaho, I. Valalik, K. Vicsi, Parkinson's disease severity estimation on hungarian speech using various speech tasks, in: 2019 10th Int. Conf. Speech Technol. Human-Computer Dialogue, SpeD 2019, 2019. https://doi.org/10.1109/SPED.2019.8906277.

[49] O. Yaman, F. Ertam, T. Tuncer, Automated Parkinson's disease recognition based on statistical pooling method using acoustic features, Med. Hypotheses. 135 (2020). https://doi.org/10.1016/j.mehy.2019.109483.

[50] H. Kuresan, D. Samiappan, S. Masunda, Fusion of wpt and mfcc feature extraction in parkinsons disease diagnosis, Technol. Heal. Care. 27 (2019) 363-372. https://doi.org/10.3233/THC-181306.